\documentclass{article} % For LaTeX2e
\usepackage[preprint]{colm2024_conference}
\pdfoutput=1

\usepackage{microtype}
\usepackage{hyperref}
\usepackage{url}
\usepackage{booktabs}
\usepackage{graphicx}
\usepackage{mathtools}
\usepackage{array,multirow}
\definecolor{darkblue}{rgb}{0, 0, 0.5}
\hypersetup{colorlinks=true, citecolor=darkblue, linkcolor=darkblue, urlcolor=darkblue}

\DeclarePairedDelimiter\ceil{\lceil}{\rceil}

\usepackage{algorithm}
\usepackage{algpseudocode}
\usepackage{tcolorbox}
\usepackage{pifont}
\usepackage{wrapfig}
\usepackage{lineno}
\usepackage{amsmath,amssymb,amsfonts}
\usepackage{algorithm}
\usepackage{algorithmicx} % <===========================================
\usepackage{algpseudocode} % <==========================================

\usepackage{xcolor} % <==========================================

\newcommand{\cmark}{\color{blue}{\ding{51}}}%
\newcommand{\xmark}{\color{red}{\ding{55}}}%

\mathchardef\mhyphen="2D

\title{
\sys{}:\\
Interpretable Tree-Adaptive Grounded Reasoning}

\author{Kate Sanders \\
Department of Computer Science\\
Johns Hopkins University\\
\texttt{ksande25@jhu.edu} \\
\And
Benjamin Van Durme \\
Department of Computer Science\\
Johns Hopkins University\\
\texttt{vandurme@jhu.edu} 
}

\newcommand{\sys}{\textsc{Bonsai}}

\usepackage{mdframed}
\usepackage{lipsum,framed}

\newcommand{\prompt}[2]{
        \subsubsection{Prompt: {#1}}
        {#2}\\~\\
        \textbf{--END PROMPT--}
        \\
    }

\begin{document}
%\linenumbers

\maketitle

\begin{abstract}
\label{sec:abs}

To develop general-purpose collaborative agents, humans need reliable AI systems that can (1) adapt to new domains and (2) transparently reason with uncertainty to allow for verification and correction. Black-box models demonstrate powerful data processing abilities but do not satisfy these criteria due to their opaqueness, domain specificity, and lack of uncertainty awareness. We introduce \sys{}, a compositional and probabilistic reasoning system that generates adaptable inference trees by retrieving relevant grounding evidence and using it to compute likelihoods of sub-claims derived from broader natural language inferences. \sys{}'s reasoning power is tunable at test-time via evidence scaling and it demonstrates reliable handling of varied domains including transcripts, photographs, videos, audio, and databases. Question-answering and human alignment experiments demonstrate that \sys{} matches the performance of domain-specific black-box methods while generating interpretable, grounded, and uncertainty-aware reasoning traces.
\end{abstract}
\section{Introduction}
\label{sec:intro}
Human professionals often write full documents describing the veracity and scope of individual claims, while many AI systems consider them to simply be true or false. To be useful in practical settings, reasoning systems must be able to model concerns like subjectivity, epistemic uncertainty, and ambiguity in natural language statements, and should be able to identify which portions of the statements these concerns apply to. Furthermore, these systems should be robust to knowledge sources of different modalities, as in many settings a grounding source may be a research report, a photograph, or a news article with embedded video clips. With these ideas in mind we introduce \sys{}, an adaptable reasoning tree generator for transparent, grounded, and probabilistic inference. \sys{} introduces a set of key design choices that enable sophisticated interpretable reasoning.

First, \sys{} extends the ``evidence extraction" paradigm -- in which natural language summaries of complex or out-of-distribution source documents are generated as data to reason over~\citep{li2024minimalevidencegroupidentification} -- to multimodal content. It accomplishes this by mapping non-textual data to evidence banks of natural language observations which it draws from during reasoning. \sys{} applies contextual conditioning to generated observations to mitigate the well-documented issue of perspective and attention ambiguity in multimodal domains~\citep{zur2024updatingclippreferdescriptions}.

\sys{} accounts for uncertainty and subjectivity in data by replacing categorical labels used in traditional claim verification (``true", ``false", etc.) with scalar likelihood scores. We introduce an iterative approach to scalar likelihood score calculation using retrieved evidence samples as explanatory conditional variables, inspired by \citet{doi:10.1126/science.185.4157.1124}'s ``anchoring and adjustment" framework for human judgments. This approach integrates naturally with chain-of-thought~\citep{wei2022chain} and allows \sys{} to behave as an \textit{evidence-grounded} adaptable prediction system~\citep{mohri2024language, jiang2025conformal} that can restructure its output depending on risk threshold.

\begin{figure}
\includegraphics[width=\linewidth]{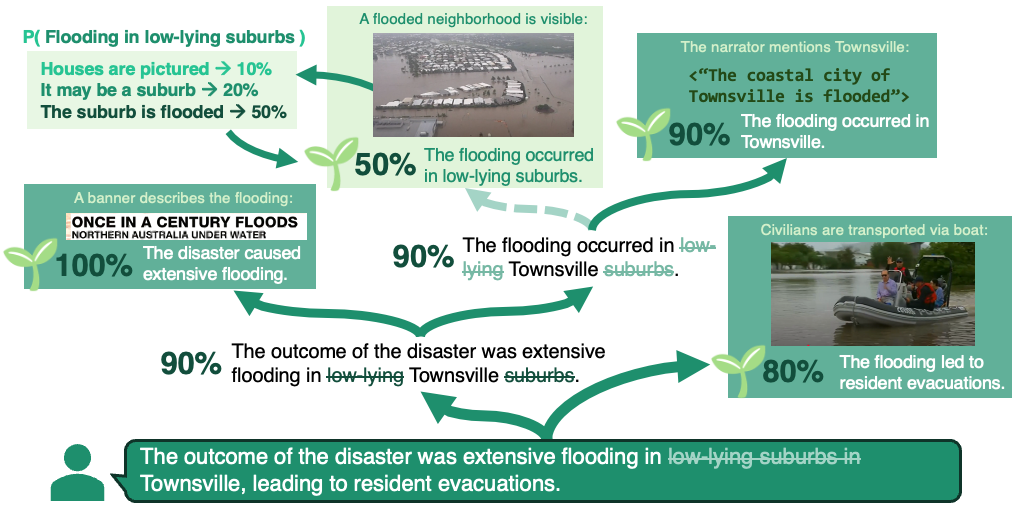}
\vspace{-10pt}
\caption{A reasoning tree over a news video as a grounding source. \sys{} recursively decomposes natural language statements about data into small, verifiable pieces. It uses retrieved evidence samples from multimodal knowledge sources to iteratively score these pieces in terms of how likely each piece is. This procedure results in grounded likelihood scores for leaves of compositional tree structures representing the original claim, alongside natural language explanations. Low-scoring branches of a reasoning tree may then be pruned, shown by the strikethrough text in the original statement and sub-claims.}
\vspace{-10pt}
\label{fig:page1}
\end{figure}

Finally, \sys{} outputs grounded, natural language explanations for each sub-claim judgment, which may be propagated upwards to generate likelihood scores and explanations for any set of sub-claims. Since \sys{} decomposes claims into tree structures (shown in Figure~\ref{fig:page1}) this enables straightforward human analysis and quick correction of intermediate sub-claims, in contrast to individually assessing every atomic claim, one-by-one. Using tree structures, contextualized evidence extraction may also be computed at arbitrary levels of granularity, and through this we adapt test-time search scaling~\citep{zhao2025sample} to facilitate flexibility in performance-compute tradeoffs with test-time \textit{evidence} search scaling. 

In summary, \sys{} is a transparent and probabilistic multimodal reasoning system that promotes three key ideas: (1) Mapping raw data to contextualized natural language observations enables high-performance modality-agnostic reasoning,
(2) assigning sub-claims with probabilistic scalars enables adaptable prediction and nuanced reasoning over ambiguities, and
(3) coupled with (1) and (2), a tree-based decomposition structure can enable impactful compute scaling and easy human-in-the-loop interpretation and corrections. Through experiments, we demonstrate that \sys{} enables state-of-the-art performance on both single- and multi-modal tasks such as EntailmentBank~\citep{dalvi2021explaining} and TVQA~\citep{lei2018tvqa}, while crucially providing a fully grounded, human interpretable thought process.
\section{Related Work}
\label{sec:rel_work}

\subsection{Probabilistic Reasoning with LLMs}
Earlier research in probability estimation has explored fine-tuning models on human probability judgments~\citep{chen2019uncertain}.
Recent work has investigated if LLMs can perform probabilistic calculations, such as estimating percentiles and basic probability computations~\citep{paruchuri2024odds}, and even update belief states given new information~\citep{qiu2024can}. The BIRD framework leverages LLM judgments for downstream probability inference~\citep{feng2024bird}, and \citet{piriyakulkij2024doing} and~\citet{mo2024tree} leverage Monte-Carlo algorithms. Notably, some approaches consider graph-centric probabilistic reasoning, either using explicit knowledge graphs~\citep{li2024enhanced} or graphs extracted via chain-of-thought reasoning~\citep{razghandi2025cer}. A growing body of work considers LM uncertainty quantification~\citep{xiao2022uncertainty, stengel2024lacie, chen2023quantifying, wang2022self}, some directly via prompting~\citep{tian2023just, xiong2023can, lin2022teaching}. Some methods incorporate conformal prediction by considering object sets in their probabilistic reasoning~\citep{ozturkler2022thinksum}.
%\subsection{Conformal Prediction}
Conformal prediction produces answer sets or intervals with assigned correctness probability scores~\citep{angelopoulos2023conformal}. Many applications to language modeling have been identified in recent years: \citet{quach2023conformal} and \citet{cherian2024large} propose a framework for conformal language modeling, \citet{mohri2024language} leverage conformal prediction for correctness guarantees, and \citet{jiang2025conformal} introduce a pragmatics-inspired framework for factuality and specificity tradeoffs.% Meanwhile, other frameworks have been introduced for structured text prediction~\citep{zhang2024conformal}, question-answering~\citep{li2023traq}, and hierarchical classification~\citep{mortier2022set}. \citet{wang2024sample} and \citet{li2025towards} apply these ideas to VLMs, to combat hallucination of textual outputs to image- and video-based inputs.

\subsection{Transparent Reasoning}
While chain-of-thought~\citep{wei2022chain} and related approaches~\citep{besta2024graph, yao2023tree, xia2024beyond} are critical to understanding LLM explanations, alongside the complementary vein of research in reasoning model traces~\citep{jaech2024openai, guo2025deepseek, muennighoff2025s1}, these methods generally lack the trustworthiness of fully transparent reasoning approaches~\citep{lanham2023measuring, yeo2024interpretable, bentham2024chain}. Such lines of work include entailment tree generation, in which claims are recursively decomposed and verified through entailment using an underlying knowledge source~\citep{weir2022nellie}, Proof of Thought, a first-order logic-inspired LLM approach~\citep{ganguly2024proof}, other tree- or graph-based reasoning methods~\citep{luo2023reasoning, mei2024inductive} such as those leveraging Monte-Carlo Tree Search~\citep{gao2024interpretable}, and other approaches that draw more direct inspiration from chain-of-thought~\citep{lyu2023faithful, chen2024visual}.

\subsection{Claim Verification}
%There is a long line of work addressing fact-checking and claim verification~\citep{bekoulis2021review, guo2022survey}. 
LLMs have enabled significant progress in the field of fact-checking~\citep{bekoulis2021review, dmonte2024claim} on a variety of benchmarks including SciFact~\citep{wadden-etal-2020-fact}, FEVER~\citep{thorne2018fact}, FactScore~\citep{min2023factscore}, and X-FACT~\cite{gupta2021xfactnewbenchmarkdataset}. Many approaches~\citep{chen2024complexclaimverificationevidence} center on claim decomposition~\citep{wanner2024closerlookclaimdecomposition}, and retrieval-augmented generation~\citep{gao2024retrievalaugmentedgenerationlargelanguage} is often applied in claim verification settings, sometimes with notable success~\citep{xu2024searchinthechaininteractivelyenhancinglarge, kao-yen-2024-magic}. Recent work has addressed evidence extraction as an intermediate step in claim verification~\citep{cao2024largelanguagemodelsdetect, li2024minimalevidencegroupidentification}. Other work focuses on accurate attributions of claims generated by the models themselves~\citep{press2024citemelanguagemodelsaccurately, weller2024accordingpromptinglanguage}. \citet{srikanth2025nli} apply an iterative inference procedure over decomposed evidence to traditional and defeasible NLI. ClaimVer~\citep{dammu2024claimver} grounds decomposed claims to source documents with generated explanations. Our work additionally incorporates multimodality and human-interpretable, sub-claim-level probability scores and explanations.
\section{\sys{} Reasoning Tree Generation}
A \sys{} reasoning tree begins with a single natural language statement that serves as the ``root", which is recursively decomposed into (usually binary) sub-claims until the claims reach an atomic state. These decompositions are included in the trace and serve as the branches of the tree. A leaf is made up of an atomic sub-claim paired with (1) the top-k most relevant evidence pieces from the grounding data (documents, videos, databases, etc.), (2) a sub-claim likelihood score, and (3) a natural language explanation detailing how the evidence was used to compute that score. These scores and explanations may be propagated up the tree branches, which we explore in Section~\ref{sec:inf}. Below, we provide further details regarding the remaining aspects of tree construction.

\subsection{Claim Decomposition}\label{subsec/decomposition}

\sys{} produces a hierarchical representation of individual claims, similar to the structure of an incomplete entailment tree. We begin with the decomposition of the initial hypothesis into compositionally entailing premises. We use GPT-4o~\citep{hurst2024gpt} to compute decompositions and provide one example decomposition for guidance, specifying the syntactic nature of the decompositions. This repeats until individual sub-claims reach an atomic state (in this setting, no longer syntactically decomposable) or the depth of the tree reaches limit $k$.  Although it is not encouraged, occasionally GPT outputs more than two sub-hypotheses for a given decomposition. We allow and account for this behavior as it generally occurs in scenarios where $>2$ premises is appropriate, e.g., ``The hurricane affected Barbuda, Cuba, and Haiti" $\rightarrow$ ``The hurricane affected Barbuda" $+$ ``The hurricane affected Cuba" $+$ ``The hurricane affected Haiti".

\subsection{Evidence Extraction and Retrieval} \label{subsec/evidence_extraction}
\sys{} performs reasoning over prespecified single- or multimodal grounding sources. To enable robust modality- and domain-agnostic reasoning, \sys{} constructs an evidence bank of natural language observations derived from the grounding source instead of using domain-specific representations. These observations are individually mapped to specific spans of the grounding source, enabling \sys{} reasoning to be fully grounded. These spans are determined offline and uniformly: Text is split into partially-overlapping windows of 6 to 12 lines depending on length. Images are passed in individually with no preprocessing. Between 1 and 10 frames are sampled from videos, depending on length. ASR is performed on audio content, and the extracted text is partitioned as a regular text document. We prompt LLMs and MLLMs to extract these observations, asking for a set of captions over small samples of the source data to ground observations in specific portions of the grounding source. These prompts may include context to improve reasoning ability (discussed further in Section~\ref{subsec/scaling}), and individual prompts for are included in Appendix~\ref{a:p}.

\begin{figure}
\includegraphics[width=\linewidth]{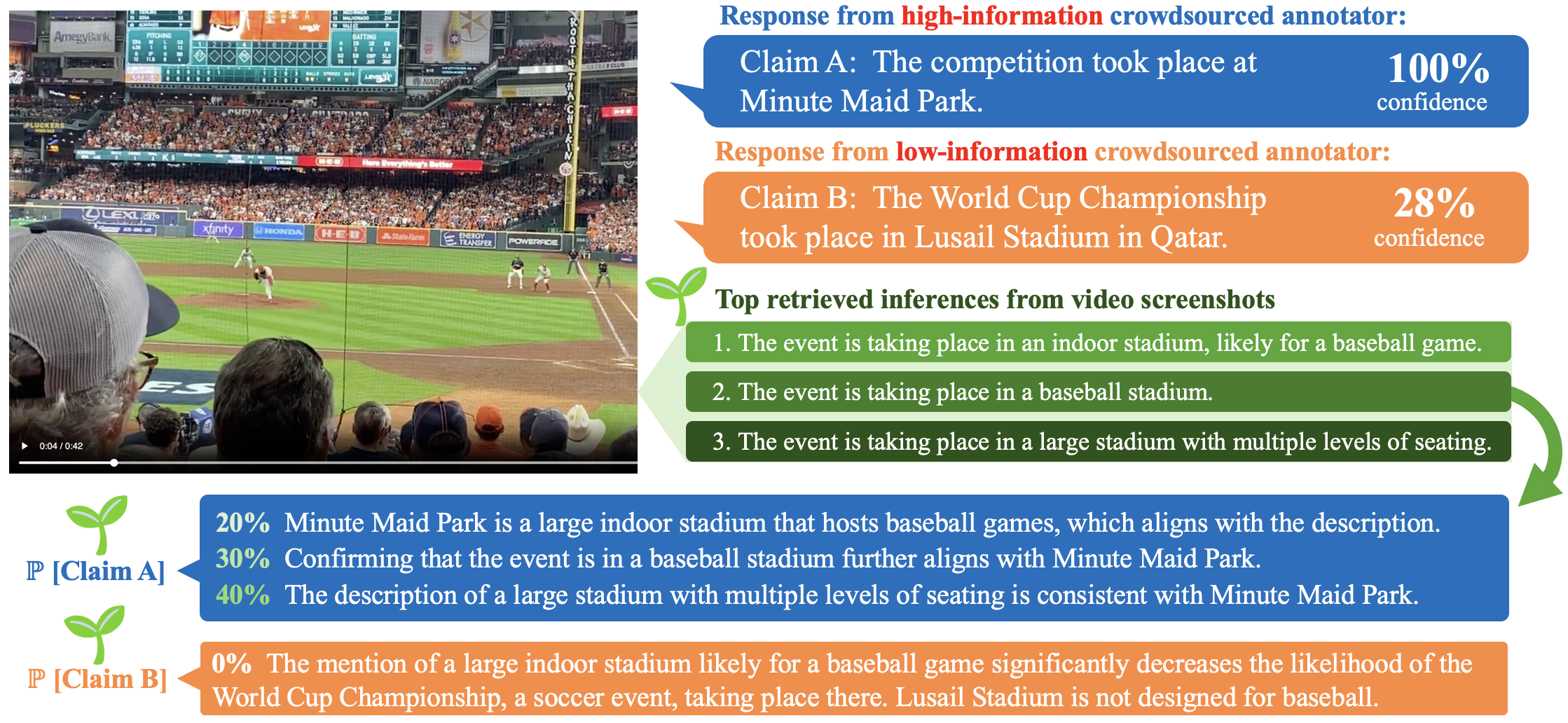}
\vspace{-10pt}
\caption{For most tasks, human performance varies depending on background knowledge. Two humans were given a set of candidate descriptions for the video on the left, and pictured on the right are their answers and confidence scores. Illustrated below the responses, \sys{} retrieves its top three generated video observations and uses them to score these claims, in the positive case by iteratively updating their likelihood scores and providing explanations.}
\vspace{-10pt}
\label{fig:page2}
\end{figure}

Retrieval over this generated evidence bank is necessary, as it becomes intractable to consider all evidence factors in most scenarios when computing the likelihood of a sub-claim. So, we use a heuristic to identify the most promising evidence factors. We pass a set of pre-extracted natural language factors into a cross-encoder model trained on the MS-MARCO dataset~\citep{bajaj2016ms} to identify the top-k evidence pieces. Increasing $k$ generally improves model reasoning at the cost of compute. $k=3$ to $k=10$ are often practical.

%\subsection{Enhancing Temporal Reasoning}
When working with temporal data, this evidence extraction approach notably eliminates temporal (or other ordering) information, which may be critical in applications like video understanding benchmarks. To account for this, we explore the application of an adjustment in which we prepend approximate temporal metadata to retrieved evidence snippets, order the evidence presented to the probability scorer temporally, and include a line to the probability scorer indicating that temporal information may be present.

\subsection{Sub-Claim Scoring}

Given some sub-claim and a set of observational evidence factors $\{f_e^{(0)},...,f_e^{(n-1)}\}$, we leverage LLM agents as ``knowledge experts" from which we elicit probabilistic judgments about the sub-claim. Following research in economics~\citep{doi:10.1126/science.185.4157.1124}, we frame this problem as an ``anchoring and adjustment" task through which we leverage LLMs' demonstrably strong chain-of-thought reasoning.

We provide an LLM with an initial guidance rubric in which we carefully describe the probabilistic scale from 0 to 100 in human-interpretable terms. We provide the agent with an artificially constructed evidence factor, or initial premise, and elicit an initial ``anchoring" probability score following the rubric. The artificial factor or premise may be constructed by generating a brief summary of the input observations. This produces an initial prior anchor score that is conditioned on a generic scenario within the scope of the specific world state. From here, the model is presented with the real evidence factors and is instructed to iteratively adjust its initial anchor score, factor-by-factor. Sample observations are shown in Figure 2, compared against two real human assessments of a video.
\section{Inference with \sys{}}
\label{sec:inf}

Given \sys{}'s detailed reasoning traces, there are multiple ways to perform inference depending on the task and compute. In this section, we consider a probabilistically sound approach to generate individual likelihoods as well as a counterfactual reasoning method for multiple choice question-answering. We also touch on a widely applicable test-time search scaling method for improved performance.

\subsection{Complete Probabilistic Inference}
\label{subsec/probability}

We may leverage the entailment tree structure of \sys{}'s reasoning traces to compute the conditional relationships between sub-claims. Let our observational evidence for sub-claim $e$ be $O_e=\{f^{(0)}_e,f^{(1)}_e,...,f^{(n-1)}_e\}$. If factors $A$ and $B$ syntactically compose $H$, then we make the assumption $P\,(H)=P\,(A\,\cap\,B)$, and consequently, $P(H|O)=P(A|B,O)\;P(B|O)=P(B|A,O)\;P(A|O)$. As evidence factors and sub-claims are both natural language strings, we may simply view the computation of $P(A|B,O)$ as $P(A|\{f^{(0)}_A,f^{(1)}_A,...,f^{(n-1)}_A,B\})$, with no modifications to the likelihood computation method.

Letting $A$ entail $(C,D)$, then this propagation may operate recursively as $P(H|O)=P(C|D,B,O)P(D|B,O)P(B|O)$. The remaining issue is how to select which variable to condition on ($A$ vs. $B$, or $C$ vs. $D$) in such a decomposition. Ideally, both would be computed and expert aggregation would be performed, but this approaches exponential complexity as the structure of the decompositional tree grows. Therefore, we only compute either $P(A|B)$ or $P(B|A)$. This full process is detailed in Algorithm~\ref{alg:generate}. A sample \sys{} output is in Appendix~\ref{a:ex}.

\begin{algorithm}[t!]
\caption{Probabilistic inference, $\textproc{Infer}$}
\begin{algorithmic}[1]
\Require Decomposition tree root $r$, evidence factor set $\mathcal{F}$ (the set of all observational evidence pieces from the grounding source), additional conditional factors $\mathcal{F}_c$ (from other propagated branches), and anchor factor $f_a$.
\Ensure Likelihood of root factor $r$ conditioned on evidence, $P(r|\mathcal{F},\mathcal{F}_c,f_a)$.\vspace{10pt}
\If{$\textproc{Children}\;(r) = \emptyset$} \Comment{Check if the current root is a leaf of the tree.}
    \State $\mathcal{F}' \gets \mathcal{F}_c\cup\textproc{Retrieve}_\varepsilon\;(r, \mathcal{F})$  \Comment{Combine evidence and propagated factors.}
    \State $P_0\gets\textproc{Anchor}\;(f_t,f_a)$ \Comment{Compute base probability score.}
    \State $\mathcal{F}^+\gets \emptyset$
    \For {$f'\in\mathcal{F}'$}
        \State $\mathcal{F}^+\gets \mathcal{F}^+\cup\{f'\}$
        \State $P_0\gets \textproc{Adjust}\;(P_0,r,\mathcal{F}^+)$ \Comment{Update probability for each piece of evidence.}
    \EndFor
\Else
    \State $A,B\gets \textproc{Children}\;(r)$
    \State $P_A\gets \textproc{Infer}\;(A,\mathcal{F}',\mathcal{F}_c\cup B,f_a)$ \Comment{Recurse on child branches, propagate factors.}
    \State $P_B\gets \textproc{Infer}\;(B,\mathcal{F}',\mathcal{F}_c,f_a)$
    \State $P_0\gets P_A\cdot P_B$ \Comment{Compute root probability.}
\EndIf
\State \textbf{return} $P_0$
\end{algorithmic}
\label{alg:generate}
\end{algorithm}

\subsection{Counter-Factual Reasoning}
\label{subsec/counterfactual}
Results show that sampling multiple answers and reasoning over them can result in higher performance on tasks than direct inference with complex reasoning models~\citep{zhao2025sample}. Similarly, many domains directly facilitate the comparison of multiple options, such as multiple choice QA. In such scenarios where multiple options are being considered, we introduce two primary additions to \sys{} to enable strong counterfactual reasoning. (1) In many cases, these different sampled answers may share similar components of their respective claim decomposition trees. We prune leaf nodes of answer trees that are inherently entailed by the other hypotheses using a cross encoder trained on SNLI~\citep{bowman2015large}, so that when making a decision, we consider the primary factors that distinguish the options from one another. (2) In cases where we know one of the inferences are true (for example, multiple choice), we can provide this information as conditional context to limit the world space being considered by the probability scoring system: Instead of measuring the general probability of a hypothesis, we can directly compute the relative likelihoods of different options conditioned on the fact that one must be true.

Given a set of counterfactual reasoning trees, there are multiple methods to select a final answer. While most probabilistically sound, we find that constructing final probability scores using the process detailed in Section~\ref{subsec/probability} (or similar approaches) penalizes more complex answers with more leaves. Therefore, in practical comparison settings we opt to take the average score across all leaves. However, this suffers from not modeling any interdependencies between leaves, as well as not sufficiently penalizing scores that accrue one or more low-probability leaves. To remedy these issues (at the cost of transparency) we also consider an LLM-based ``judge" method that outputs a final answer based on the collection of leaf sub-claims and their likelihood computed scores.

\subsection{Test-Time Evidence Search Scaling}
\label{subsec/scaling}

Depending on the complexity of the underlying data, the generic evidence extracted through \ref{subsec/evidence_extraction} may not be sufficient to output a high probability score for any complex claims about the content. In such a case, this will be demonstrated by the consistently low resulting probability scores in a multiple-choice or multi-inference setting. Given these scores, the system may choose to engage in a second (or $n$th) round of evidence extraction and claim re-scoring. In this re-extraction, claims or sub-claims constructed during the claim decomposition step (\ref{subsec/decomposition}) may be passed in as contextualizing information to the extractor, resulting in more specific and topical observations. While increasingly computationally expensive, as the context used for evidence extraction grows more specific the confidence of the model correspondingly improves. We illustrate the benefit of such an approach in Section~\ref{subsec/multivent}.
\section{Experiments}
We evaluate \sys{} on four tasks. We first consider the quality of the proposed calibration approach, using a dataset of human-scored ambiguous images (Section~\ref{subsec:squide}). Then, we test the full reasoning system on traditional single- and multi-modal question-answering tasks (Section~\ref{subsec/mcqa}). We finally evaluate the system on a multimodal inference task that involves reasoning over ambiguity (Section~\ref{subsec/multivent}). Full setup information is included in Appendix~\ref{a:es}.

\begin{wrapfigure}{L}{0.5\textwidth}
    \centering
    \vspace{-15pt}
    \includegraphics[width=.5\textwidth]{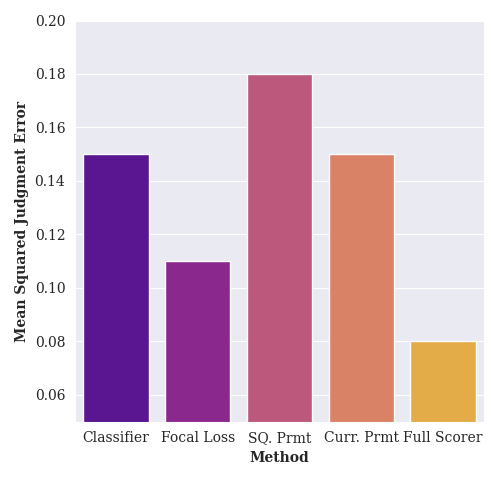}
    \vspace{-20pt}
    \caption{Agreement of different uncertainty quantification approaches compared to human likelihood judgments on a visual classification task. Unlike basic probability scoring prompts (SQ. Prmt and Curr. Prmt), \sys{}'s ``anchor and adjust" evidence-focused approach (Full Scorer) outperforms traditional uncertainty quantification methods with a fine-tuned classifier (Focal Loss).}
    \label{figure/squid}
    \vspace{-30pt}
\end{wrapfigure}

\subsection{Likelihood Calibration}
\label{subsec:squide}
We first characterize the quality of probability judgments produced by \sys{} by comparing them against human probability judgments in a low-information setting.

\paragraph{Task} We evaluate \sys{} on SQUID-E~\citep{sanders2022ambiguous}, a collection of ambiguous images sampled from videos of everyday events like weddings and medical procedures. Alongside ground truth labels of these events, the images are labeled with human probability judgments on a 0-100 scale that quantify how likely people believe the images depict these ground-truth events. We evaluate on the full set of 3,600 human-labeled images. We evaluate the MSE of \sys{}'s probability judgments and the median human judgments for each image in the test set. We compare \sys{} against a traditional visual event classification network, this network combined with the most effective model calibration technique identified in the SQUID-E publication, focal loss~\citep{lin2017focal}, alongside alternative LLM-driven probability solicitation approaches.

\paragraph{Setup} We use Molmo 7B~\citep{deitke2024molmo} for image evidence extraction, and GPT-4o for probability scoring. As the hypotheses in this task are already atomic (essentially the one- or two-word description of the event) we do not further decompose the hypothesis. We test two prompts alongside the traditional classification approaches (Classifier, Focal Loss) and \sys{}'s probability scoring algorithm (Full Scorer): SQ. Prmt is a prompt-adapted version of the original instructions given to human annotators for SQUID-E, and Curr. Prmt is a prompt-adapted version of the probability scoring outlines used in \sys{}, but without the anchoring-and-adjustment portion used with extracted visual evidence.

\paragraph{Results} Agreement against ground-truth human scores is shown in Fig.~\ref{figure/squid}. While the likelihood rubric used in \sys{} on its own outperforms a basic likelihood scoring prompt and matches the alignment of fine-tuned classifier logits, it underperforms compared to the uncertainty quantification approach. Only \sys{}'s anchor-and-adjust method aligns with humans better than all other methods, indicating the efficacy of a more in-depth probability scoring method when considering human alignment.

\subsection{Traditional Multiple Choice QA}
\label{subsec/mcqa}
We explore whether \sys{} maintains high-quality performance on traditional question-answering tasks in text and multimodal domains. 

\paragraph{Tasks} Following the experimental setup of existing reasoning trace generation work, we evaluate on EntailmentBank~\citep{dalvi2021explaining} and TVQA~\citep{lei2018tvqa}. EntailmentBank consists of multiple choice questions taken from middle school science curricula (from the ARC dataset~\citep{clark2018think}), and provides an evidence bank of factual science statements from the WorldTree dataset~\citep{jansen2018worldtree} to use as supporting evidence for answers. TVQA consists of multiple choice questions about popular TV shows, and provides episode clips and dialogue transcripts as supporting evidence to draw from. We sample 1,000 questions from each dataset (requiring sampling from the training set for EntailmentBank---scores from individual train/test splits are included in Appendix~\ref{a:er}). For both datasets we use mean accuracy as the evaluation metric, and we compare \sys{} against contemporary end-to-end models designed for the tasks as well as architecturally similar neuro-symbolic methods. For EntailmentBank, there is not a naturally fair black-box comparison to the transparent systems, as the WorldTree corpus includes 11,941 natural language facts, or over ~150K tokens. To approximate, we run GPT on the standard ARC benchmark, in which they use their parametric knowledge instead of a grounding source.

\paragraph{Setup} For both datasets, we use GPT-4o to convert each answer choice into a natural language hypothesis. For example, the question answer pair ``Q: At what temperature does ice melt? A: 72 degrees Celsius" would become ``Ice melts at a temperature of 72 degrees Celsius". For EntailmentBank, we use the provided WorldTree factbase as our precomputed evidence collection. For transcript text inference on TVQA, we use GPT-4o with context windows of 6 lines of dialogue, and for vision we use Molmo 7B with approx. 10 sampled frames per video. For conditioning context for probability scoring in EntailmentBank we use the original question with the framing ``someone is asking the question, ...", and for TVQA we extract a short transcript summary with GPT-4o. For both tasks we use the counterfactual reasoning approach detailed in Sec.~\ref{subsec/counterfactual}. We use the two scoring alternatives in Section~\ref{subsec/counterfactual}: Aggregation of leaves (via multiplication for ARC and mean for TVQA, as ARC has similar leaf counts per answer while TVQA does not), and a GPT-4o judge that takes leaves and outputs the likeliest answer (written \sys{}$_{J\mhyphen4o}$).

\begin{table}
    \centering
    \begin{tabular}{cccccc}
    \toprule
        \textbf{Model} & \textbf{Transp.} & \textbf{EB/ARC Acc.} & \textbf{Model} & \textbf{Transp.} & \textbf{TVQA Acc.}\\
        \midrule
        NELLIE & \cmark & 71.4 & VideoChat2 & \xmark & 40.6 \\
        TreeWise & \cmark & 79.2 & TV-TREES & \cmark &  49.4 \\
        Entailer-11B & \xmark & 74.1 & VideoChat-Mistral & \xmark & 50.6 \\
        {\color{gray}GPT-3.5*} & \xmark & {\color{gray}88.7} & MiniGPT4-Video & \xmark &  54.2 \\
        {\color{gray}GPT-4o*} & \xmark & {\color{gray}96.0} & IG-VLM & \xmark & 57.8\\
        \midrule
         \textbf{\sys{}}  & \cmark & 87.7 & \textbf{\sys{}} & \cmark & 65.5\\
         \textbf{\sys{}$_{J\mhyphen4o}$}  & \cmark & \textbf{95.6} & \textbf{\sys{}$_{J\mhyphen4o}$} & \cmark & \textbf{68.8}\\
        \bottomrule
    \end{tabular}
            \caption{\sys{} performance on traditional multiple-choice tasks, EntailmentBank and TVQA, compared against other transparent and black-box methods (all zero-shot). GPT-3.5 and GPT-4o are run on standard ARC(*), not EntailmentBank with the WorldTree grounding source. On both tasks, \sys{} with basic scoring already outperforms other transparent approaches, and using a 4o judge (\sys$_{J\mhyphen4o}$) it achieves another performance boost.}
    \vspace{-10pt}
\end{table}

\paragraph{Results} 
On EntailmentBank, the raw \sys{} scorer outperforms related transparent approaches. This illustrates that while \sys{} can generalize beyond text-centric tasks, it can still retain appropriate performance on more traditional text benchmarks. It should be noted, however, that where \sys{} improves in generalizability to ambiguity, it loses in its ability to systematically prove entailment: In some scenarios, such as a task like EntailmentBank, guaranteed entailment ensured by an approach like TreeWise~\citep{weir2024enhancing} may be a superior output than calibrated probabilities without entailment guarantee. On TVQA, the raw \sys{} scorer also outperforms other methods developed for the benchmark, indicating a strong ability to reason over multiple modalities simultaneously.

The LLM-assisted probability aggregation is effective on both tasks. This is reasonable, as the alternate methods of modeling probabilities either penalize the system for comprehensive decompositions in the case of the Sec.~\ref{subsec/probability} approach, or fail to represent conditional dependencies in the case of the leaf score aggregation method used in these experiments. It is likely that the LLM judge ad-hoc models the dependencies while implicitly conditioning over tree size, resulting in a balanced aggregation approach.

\begin{wraptable}{r}{0.5\textwidth}
    \centering
    \vspace{-30pt}
    \begin{tabular}{ccc}
    \toprule
        \textbf{Model} & \textbf{Transp.} & \textbf{MV Acc.} \\
        \midrule
        LLaVA-Next-7B & \xmark & 61 \\
        InternVL 2.5-8B & \xmark & 83\\
        Qwen VL 2-7B & \xmark & 87\\
        \midrule
        \textbf{\sys{}} & \cmark & 76\\
    %    \textbf{\sys{}$_{SR}$} & \cmark & \_\\
        \textbf{\sys{}$_{SL}$} & \cmark & \textbf{89}\\
    %    \textbf{\sys{}$_{SL}$} & \cmark & \textbf{90}\\
        \midrule
        Human (Avg.) & \cmark & 93\\
        Human (Best) & \cmark & 100\\
        \bottomrule
    \end{tabular}
    \label{tab:multivent}
        \caption{Results of the ambiguous video understanding task against SoTA video models. \sys{}'s results demonstrate the power of evidence scaling, matching the performance of SoTA black-box video models.}
    \vspace{-5pt}
\end{wraptable}

\subsection{Ambiguous Video Analysis}
\label{subsec/multivent}

\paragraph{Task} We evaluate \sys{} on MultiVENT~\citep{sanders2023multivent}, a dataset of short to long-form videos depicting portions, but not the entirety, of various real-world current events. We scrape English news articles describing the events depicted in the dataset and use GPT-4o to generate various factual statements about these events. We then sample a set of 46 English document-video pairs using the MultiVENT 1.0 release, cluster a set of five statements about distractor events that are most semantically similar to the correct event statements (using a cross-encoder trained on MS-MARCO), and use these clusters as multiple-choice questions about the video content (see Appendix~\ref{a:mv} for examples. Full task will be released alongside code). We first solicit high-performing human crowdsource annotators to watch the videos and complete the QA task with two-way redundancy (see Appendix~\ref{a:i}). We evaluate \sys{}'s ability to identify the correct statement about each video, and use mean accuracy as the evaluation metric. We compare the system's performance against state-of-the-art video understanding models (with similar vision backbone size) and human performance (debatable vision backbone size). Human performance is reported via mean score and via the best individual annotator, who annotated all videos.

\paragraph{Setup} We replicate the setup described in Sec.~\ref{subsec/mcqa}. For audio, we peform ASR with Whisper~\citep{radford2022robustspeechrecognitionlargescale} and sample 6 ``sentences" per window for evidence extraction. We demonstrate the value of ``test time evidence extraction" by comparing multiple ways of sampling evidence. We first take the traditional approach used in Sec.~\ref{subsec/mcqa}, passing in a general question that each of the hypotheses attempt to answer. We compare this against re-sampling visual evidence using the leaf sub-claims as context (\sys{}$_{SL}$).

\paragraph{Results}
The results, shown in Table~\ref{tab:multivent}, illustrate \sys{}'s ability to reason over vision-centric data where audio provides limited additional information, compared to a benchmark like TVQA where dialogue-only performance can reach over 44\%~\citep{sanders2024tv}. The results also illustrate the efficacy of test-time evidence scaling, boosting \sys{} performance by 13 points and matching state-of-the-art black-box video model performance (and notably outperforming earlier 2024 models) while simultaneously providing comprehensive and grounded reasoning traces. While \sys{} does not significantly outperform Qwen-2 in this experiment, it is notable that it matches performance \textit{while producing a comprehensive reasoning trace}. Human annotators rated their answer certainty below 70\% for over a quarter of their answers on the task. %In the face of human uncertainty, there are many applications in which it is helpful to provide transparent and grounded reasoning traces to humans than individual answers.

\section{Conclusion}
We introduce a probabilistic reasoning tree generator for broadly adaptable and transparent reasoning that remains grounded in multimodal evidence. Through this system we demonstrate that probabilistic reasoning via an iterative algorithm enables robust reasoning over uncertainty that improves human alignment both for low-level and high-level tasks. Further, it demonstrates the power of leveraging contextually conditioned captions from multimodal data to enable powerful cross-modal reasoning. As a general purpose system, \sys{} trades optimized performance on specific domains for broad adaptability, and as future work we envision significant improvement across different tasks by introducing domain-specific modifications to system modules. In short, \sys{} showcases the exciting potential of transparent reasoning systems on complex, real-world challenges.
%\input{content/08_ethics}

%\subsubsection*{Acknowledgments}

\bibliography{colm2024_conference}
\bibliographystyle{colm2024_conference}

\newpage
\appendix
\section{Examples}
\label{a:ex}

Here, we provide example outputs from \sys{} on the four tasks documented in the experiments section.

\subsection{Sample Tree}

A sample tree from \sys{} on TVQA is shown in Figure 4.

\begin{figure}[!ht]
\includegraphics[width=\linewidth]{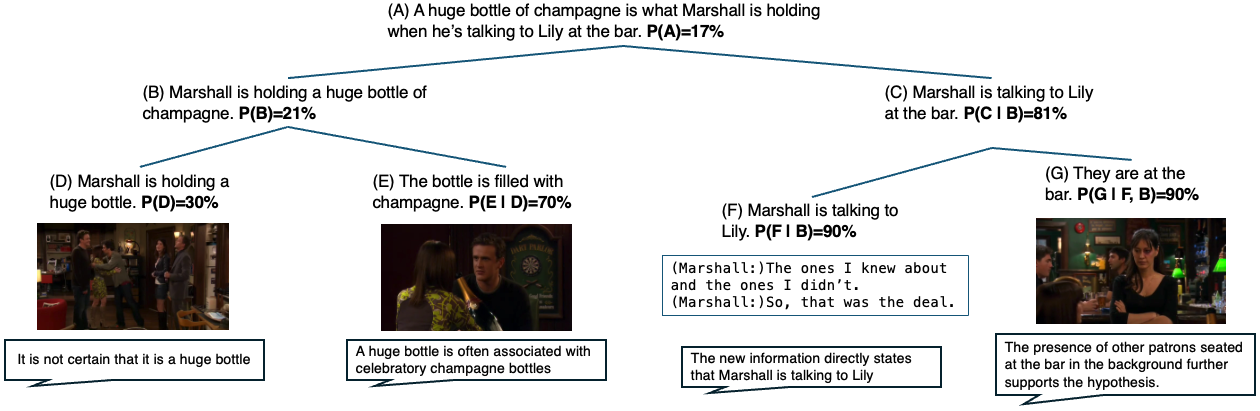}
\caption{An example \sys{} output on TVQA. One of the $\leq$3 evidence pieces per leaf and a snippet of the system's probability scoring trace per leaf are shown. Conditional probabilities are shown next to their respective sub-claims, alongside what other sub-claims they are conditioned on.}
\end{figure}

\subsection{Sample Retrieved Evidence and Probability Scores}

% 0GcyMkTt
% 4sIYaijl
% 0NorbcLU

\begin{table}[!ht]
\small
    \centering
    \begin{tabular}{p{3cm}p{3cm}p{6cm}}
    \toprule
    
        \small \textbf{Claim} & \small \textbf{Evidence} & \small \textbf{Scoring Output}\\
        \midrule
        A wedding is happening. & A man in a black suit is holding a book, suggesting a formal event or ceremony. & \textbf{30\%}: A man in a black suit holding a book suggests a formal event, which could be a wedding, but it could also be other types of ceremonies or formal gatherings.\\
        & & \\
         & A woman in a white dress is present, which is typical attire for a bride in a wedding ceremony. & \textbf{70\%}: The presence of a woman in a white dress strongly suggests a wedding, as this is traditional bridal attire. This significantly increases the likelihood of the hypothesis.\\
         & & \\
         & The presence of multiple people in formal attire and the book-holding man indicates a structured event, possibly a wedding. & \textbf{80\%}: The presence of multiple people in formal attire and the man holding a book further supports the idea of a structured event, such as a wedding. This additional context makes the hypothesis quite likely.\\
         \bottomrule
    \end{tabular}
    \caption{SQUID-E observational evidence and probability judgments.}
\end{table}

% AKDE&ED_2012_8_43.json
% Mercury_7083948.json

\begin{table}
\small
    \centering
    \begin{tabular}{p{3cm}p{3cm}p{6cm}}
    \toprule
        \small \textbf{Claim} & \small \textbf{Evidence} & \small \textbf{Scoring Output}\\
        \midrule
         Wings help set birds apart from other vertebrates. & An example of an adaptation for flight in birds is feathers; wings; hollow bones & \textbf{50\%}: This information highlights that wings are an adaptation for flight in birds, which supports the idea that wings are a significant trait for birds. However, it does not directly address the exclusivity of wings to birds among vertebrates.\\
         &  & \\
         & all species of birds have wings & \textbf{60\%}: The fact that all species of birds have wings strengthens the argument that wings are a defining characteristic of birds. This increases the likelihood that wings help set birds apart from other vertebrates.\\
         &  & \\
         & mammals and reptiles and amphibians and birds are vertebrates & \textbf{60\%}: This information confirms that birds are among the vertebrates, but it does not provide new information about the uniqueness of wings to birds.\\
         \midrule
         The order of the elements in the periodic table is determined by their abundance in nature. & elements on the right side of the periodic table form ions by gaining electrons & \textbf{10\%}: This information is about the chemical behavior of elements, not their order. It does not support the hypothesis.\\
         &  & \\
         & the Periodic Table of Elements can be used for  predicting; determining the properties of an element & \textbf{10\%}: The periodic table's use in predicting properties does not relate to the hypothesis about ordering by abundance. It does not support the hypothesis.\\
         &  & \\
         & the atomic number of elements increases by 1 from left to right on a row on the periodic table & \textbf{0\%}: This information directly contradicts the hypothesis, as it states that the atomic number increases by 1 from left to right, which is the actual basis for the order of elements.\\
         \bottomrule
    \end{tabular}
    \caption{Excerpts from EntailmentBank probabilistic reasoning traces.}
\end{table}

%122053.json
%122073.json
\begin{table}
\small
    \centering
    \begin{tabular}{p{3cm}p{3cm}p{6cm}}
    \toprule
        \small \textbf{Claim} & \small \textbf{Evidence} & \small \textbf{Scoring Output}\\
        \midrule
         Kara's brain activity is slowing down. & 00:00:14 The brain scan results show unusual activity in Kara's brain. & \textbf{20\%}: The information that Kara's brain scan shows unusual activity is consistent with all five statements, including the hypothesis. This does not change the likelihood of the hypothesis being true relative to the other possibilities.\\
         &  & \\
         & 00:01:02 Cameron observes that Kara is seizing, indicating a discrepancy in the interpretation of her brain activity. & \textbf{10\%}: Cameron observing that Kara is seizing suggests that her brain activity might be more consistent with increased activity, as seizures typically involve abnormal, excessive brain activity. This makes the hypothesis of slowing down less likely.\\
         &  & \\
         & 00:01:15 Kara's brain activity is unusual because it is slowing down instead of accelerating during the seizure. & \textbf{90\%}: The information that Kara's brain activity is unusual because it is slowing down during a seizure directly supports the hypothesis. This makes it extremely likely that the hypothesis is true.\\
         \midrule
         The garment floated.& 00:00:27 Beckett describes the garment as \"invisible,\" suggesting it cannot be seen when held up. & \textbf{20\%}: Beckett's description of the garment as \"invisible\" suggests that it might have properties related to invisibility, which could be consistent with floating. However, it does not directly indicate floating, so the probability remains the same.\\
         &  & \\
         & 00:00:43 The garment is lifted from a desk surface. & \textbf{20\%}: The garment being lifted from a desk surface does not provide any new information about whether it floated after being held up. It could have been lifted manually, so this does not change the probability.\\
         &  & \\
         & 00:01:59 Beckett comments that the garment is invisible. & \textbf{20\%}: Beckett's repeated comment that the garment is invisible reinforces the idea that the garment has invisibility properties. While this might suggest something unusual about the garment, it still does not directly indicate floating. The probability remains unchanged.\\
         \bottomrule
    \end{tabular}
    \caption{TVQA scores}
\end{table}

%1364012654717792256_749.json

\begin{table}
\small
    \centering
    \begin{tabular}{p{3cm}p{3cm}p{6cm}}
    \toprule
        \small \textbf{Claim} & \small \textbf{Evidence} & \small \textbf{Scoring Output}\\
        \midrule
         The first photons will be captured on Webb's primary camera. & The Ariane 6 rocket is visible in the image. & \textbf{30\%}: The presence of the Ariane 6 rocket suggests a space-related mission, which aligns with the hypothesis about the Webb telescope. However, it does not confirm the specific purpose of capturing the first photons.\\
         &  & \\
         & The Webb telescope is shown in its deployment configuration. & \textbf{70\%}: Seeing the Webb telescope in its deployment configuration strongly suggests that the technological development is related to the telescope, increasing the likelihood that the hypothesis is true.\\
         &  & \\
         & The Webb telescope's name is spelled out on the rocket's side. & \textbf{80\%}: The Webb telescope's name on the rocket further confirms the association with the Webb telescope, making it quite likely that the purpose is to capture the first photons on Webb's primary camera.\\
         \midrule
         The Toronto Raptors played against the Golden State Warriors. & "We the North" is a slogan associated with the Toronto Raptors, indicating a Toronto location. & \textbf{30\%}: The slogan \"We the North\" is strongly associated with the Toronto Raptors, which increases the likelihood that the event involved the Raptors. However, it does not confirm the opponent or the specific game.\\
         &  & \\
         & A photographer is visible in the background. & \textbf{30\%}: The presence of a photographer is common at many sporting events and does not provide specific information about the teams or the event. Therefore, it does not significantly change the probability.\\
         &  & \\
         & The competition likely took place in California. & \textbf{60\%}: If the competition likely took place in California, this aligns with the hypothesis that the Raptors played against the Golden State Warriors, as the Warriors are based in California. This increases the likelihood of the hypothesis being true.\\
         \bottomrule
    \end{tabular}
    \caption{MultiVENT scores}
\end{table}
\section{Experimental Setups}
\label{a:es}

Details regarding the experimental setup of \sys{} across the different tasks is shown in Table 3. The video frame sampling parameters are shown in Equation 1, let $x$ be the length of the video in seconds.

% Vision backbone
% Decomp backbone
% Prob backbone
% Text sampling
% Image sampling

% Decomp max
% Evidence max
% Temporal enhancement
% Evidence level
% Aggregation

\begin{table}[!ht]
    \centering
    \begin{tabular}{ccccccccc}
    \toprule
        \textbf{Task/Method} & \textbf{VB} & \textbf{DB} & \textbf{FS}& \textbf{DM} & \textbf{EM} & \textbf{TE} & \textbf{EL} & \textbf{AG} \\
        \midrule
        \textbf{SQUID-E / B} & Molmo-7B & \ding{55}  & \ding{55} & \ding{55} & 3 & \ding{55} & base & \ding{55}\\
        \textbf{EB / B} & \ding{55} & 4o-mini  & \ding{55} & 2 & 10 & \ding{55} & leaf & mean\\
        \textbf{EB / B$_{J-4o}$} & \ding{55} & 4o-mini  & \ding{55} & 2 & 10 & \ding{55} & leaf & 4o judge\\
        \textbf{TVQA / B} & Molmo-7B & 4o-mini  & Eqn.~\ref{frame_sample} & 3 & 3 & \ding{51} & base & mean\\
        \textbf{TVQA / B$_{J-4o}$} & Molmo-7B & 4o-mini  & Eqn.~\ref{frame_sample} & 3 & 3 & \ding{51} & base & 4o judge\\
        \textbf{MV / B} & Molmo-7B & 4o  & Eqn.~\ref{frame_sample} & 3 & 3 & \ding{55} & base & mean\\
        \textbf{MV / B$_{SL}$} & Molmo-7B & 4o  & Eqn.~\ref{frame_sample} & 3 & 3 & \ding{55} & leaf & mean\\
        \bottomrule
    \end{tabular}
    \caption{Setup details of the experimental implementations of \sys{} (B). \textbf{VB}: Vision backbone, \textbf{DB}: decomposition (and hypothesis generation) backbone, \textbf{FS}: frame sampling, \textbf{DM}: decomposition max ($k$), \textbf{EM}: evidence max ($k$), \textbf{TE}: temporal enhancement, \textbf{EL}: evidence conditioning tree level, \textbf{AG}: aggregation approach.}
    \label{tab:my_label}
\end{table}

\begin{equation}
\begin{split}
F(x)= \begin{cases}  \vspace{5pt}
      k_1 & x\leq m_1 \\ \vspace{5pt}
      \ceil{k_1+(x-m_1)(\frac{(k_2-k_1)}{(m_2-m_1)})} & m_1\leq x\leq m_2 \\ \vspace{5pt}
      \ceil{k_2+(x-m_2)(\frac{(k_3-k_2)}{(m_3-m_2)})} & m_2\leq x\leq m_3 \\ 
      k_3 & m_3\leq x 
   \end{cases}\\~\\
   k_1=1,\;k_2=6,\;k_3=10,\;m_1=3,\;m_2=20,\;m_3=40
   \end{split}
    \label{frame_sample}
\end{equation}
\section{EntailmentBank Splits}
\label{a:er}

\begin{table}[!ht]
    \centering
    \begin{tabular}{ccccc}
    \toprule
         & GPT-3.5 & GPT-4o & \sys{} & \sys{}$_{J-4o}$ \\
         \midrule
        Train & 90.9 & 97.1 & 87.9 & 96.1 \\
        Test  & 84.4 & 93.8 & 86.4 & 94.7 \\
        \bottomrule
    \end{tabular}
    \caption{Train/Test splits for the EntailmentBank/ARC experiment.}
\end{table}
\section{Probability Scoring}
\label{a:ps}
\subsection{Main Prompt}
\prompt{Probability Scoring}{
You are a reasoning system that analyzes the likelihood of complex events given information about hypothetical scenarios.\\

You are given a description of a fictional scenario and a hypothesis about that scenario that may or may not be true. Given the situation, you will first score the likelihood that this hypothesis is true, on a scale from 0 to 10, using the following rubric as guidance:\\

\textbf{0 (virtually impossible):} Essentially no way the hypothesis could possibly be true, given the evidence. Less likely than being struck by lightning.

\textbf{1 (unlikely):} The hypothesis is unlikely, but definitely not impossible.

\textbf{2 (possible):} The hypothesis could be true given the evidence, but there is better chance that it is false. Less likely than drawing a card of the suit of clubs from a standard card deck.

\textbf{3 (reasonable chance):} You would not be more than mildly surprised that the hypothesis is true. About one thirds chance.

\textbf{4 (a bit less than even-odds):} Slightly below fifty-fifty probability. You would not bet more than a small sum that the hypothesis is false.

\textbf{5 (fifty-fifty):} Given the information about the situation, there is approximately equal chance that the hypothesis is true vs. the hypothesis is false. As likely as a fair coin landing on heads.

\textbf{6 (a bit more than even-odds):} Slightly above fifty-fifty probability. You would not bet more than a small sum that the hypothesis is true.

\textbf{7 (probable):} Likely, but you would still not be overly surprised if the hypothesis turned out to be false.
8 (quite likely): About as likely as \*not\* rolling a “2” with a six-sided die.

\textbf{9 (extremely likely):} Quite certain. You would bet a large amount of money on the hypothesis being true.

\textbf{10 (practically certain):} You cannot imagine a scenario in which the hypothesis is not true, given the situational evidence.\\

Then, you will be given some new information about the situation that you may use as evidence to update your probability score. You will update your probability score incrementally, one statement at a time, in order to best consider the impact of each individual piece of information. Again, your prediction should be on the 0-10 scale and use the provided scoring rubric. Before each score update, write an explanation for your adjustment.\\

Label your initial prediction with (0), and label your updated predictions with the evidence number it corresponds to. Write your enumerated explanations and probability scores, and nothing else.\\

\{Exemplars\}\\

That is the end of the examples. Now, it’s time for you to assign probabilities to a new fictional scenario:\\

ORIGINAL DESCRIPTION: \{summary\}\\

\{Counterfactual prompt, if applicable\}\\

HYPOTHESIS: \{hypothesis\}\\

NEW INFORMATION:

\{information\}\\

PROBABILITY SCORES:
}

\prompt{Probability Scoring (Exemplars)}{
Here is a first example:\\

ORIGINAL DESCRIPTION: There were puddles in the street and dark clouds hung overhead. The Mississippi flag was visible on a nearby car.\\

HYPOTHESIS: A tornado rolled through a town in Mississippi.\\

NEW INFORMATION:\\
(1) Rubble was piled by a curb.\\
(2) Dark clouds are in the sky.\\
(3) The situation was somehow related to the Mississippi Emergency Management Agency.\\

PROBABILITY SCORES:\\
(0) EXPLANATION: It is much more likely that there was just a storm. But the scenario is likely in Mississippi.\\
SCORE: 1\\
(1) EXPLANATION: The information indicates that some damage may have happened in the scenario, but there are many other reasons that rubble could be present.\\
SCORE: 2\\
(2) EXPLANATION: This information is mostly included in the original description, and so it doesn’t change the current score.\\
SCORE: 2\\
(3) EXPLANATION: If the Mississippi Emergency Management Agency was involved, it is likely that the scenario involved more than a regular storm. In Mississippi, a tornado might be the most likely reason.\\
SCORE: 6\\

Here is a second example:\\

ORIGINAL DESCRIPTION: There is a large crowd of people gathered before a lit-up stage at night.\\

HYPOTHESIS: The band Blur performed at Coachella 2024.\\

NEW INFORMATION:\\
(1) The event described is Coachella 2024.\\
(2) Large text reading “Blur” is visible on stage.\\
(3) A rock band is performing on stage.\\
(4) The band Blur is performing on stage.\\

PROBABILITY SCORES:\\
(0) EXPLANATION: The probability that the performer and event both match up with the hypothesis is low, given the number of valid possibilities.\\
SCORE: 1\\
(1) EXPLANATION: While the event must be Coachella, it is still highly unlikely that the performer happens to match the hypothesis.\\
SCORE: 1\\
(2) EXPLANATION: The text on stage makes it extremely likely that the performer is Blur.\\
SCORE: 9\\
(3) EXPLANATION: While this evidence does further improve the odds of the hypothesis, there is still a small chance that it is incorrect.\\
SCORE: 9\\
(4) EXPLANATION: The hypothesis must be true given information pieces 1 and 4.
SCORE: 10}
\subsection{SQUID-E Alternate Prompts}
\prompt{Image Calibration Alternate Prompt A}{
Rate your confidence that the image belongs to a video depicting the provided event type on a scale from 0 to 10.\\

For reference, an image should only be rated 0 if you are nearly certain that the video it belongs to does not depict the target event type. Rating an image between 1 - 4 indicates that the visual evidence in the image suggesting it belongs to the target event type is weak enough that it is likelier that the video depicts to another event type. Where on the scale you rate it depends on the strength of the visual evidence. Rating an image at 5 indicates that you feel there is an equal likelihood that the image belongs to a video of the target event type and that the image belongs to a video of a similar event type that shares some visual attributes. Ratings between 6 - 9 indicate that it is likelier that the video depicts the target event than it doesn’t. An image should only be rated 10 if you are nearly certain that the video it belongs to depicts the target event type.\\

Write your final score and nothing else.\\

EVENT: "\{question\}"\\

SCORE:
}

\prompt{Image Calibration Alternate Prompt B}{
Rate your confidence that the image belongs to a video depicting the provided event type on a scale from 0 to 10.\\

Use the following rating scale.\\

0 (virtually impossible): Essentially no way the hypothesis could possibly be true, given the evidence. Less likely than being struck by lightning.\\
1 (unlikely): The hypothesis is unlikely, but definitely not impossible.\\
2 (possible): The hypothesis could be true given the evidence, but there is better chance that it is false. Less likely than drawing a card of the suit of clubs from a standard card deck.\\
3 (reasonable chance): You would not be more than mildly surprised that the hypothesis is true. About one thirds chance.\\
4 (a bit less than even-odds): Slightly below fifty-fifty probability. You would not bet more than a small sum that the hypothesis is false.\\
5 (fifty-fifty): Given the information about the situation, there is approximately equal chance that the hypothesis is true vs. the hypothesis is false. As likely as a fair coin landing on heads.\\
6 (a bit more than even-odds): Slightly above fifty-fifty probability. You would not bet more than a small sum that the hypothesis is true.\\
7 (probable): Likely, but you would still not be overly surprised if the hypothesis turned out to be false.
8 (quite likely): About as likely as \*not\* rolling a “2” with a six-sided die.\\
9 (extremely likely): Quite certain. You would bet a large amount of money on the hypothesis being true.\\
10 (practically certain): You cannot imagine a scenario in which the hypothesis is not true, given the situational evidence.\\

Write your final score and nothing else.\\

EVENT: "\{question\}"\\

SCORE:
}
\section{Sample MultiVENT Multiple Choice Options}
\label{a:mv}

Sample A:
\begin{itemize}
\item People present at the office building, including employees of the law firm and the suspected perpetrator, were affected by the disaster.
\item About 5,900 people from 11 provinces and cities were displaced from their homes, with roughly 4,600 of them staying in temporary shelters following disaster warnings.
\item Many individuals and communities across California were affected, including Daniel Salazar of Berry Creek, who had previously escaped the Camp fire that killed his grandparents.
\item Among the 37 found individuals, 12 were students, five of whom were confirmed dead. Additionally, the bus driver also died and 16 other individuals were injured.
\item People affected by the disaster include a man in his 60s found dead in a submerged car, a man in his 40s killed by a mudslide, and hundreds of thousands faced with power outages across southwestern Japan.
\end{itemize}

Sample B:

\begin{itemize}
\item Ten teams participated in the 2019 Cricket World Cup, including England and South Africa.
\item The Astros and the Phillies participated in the competition.
\item The Toronto Raptors and the Golden State Warriors participated in the competition.
\item Numerous nations participated in the 2016 Summer Olympics, with sports teams and individual athletes competing across 42 sports and 306 medal events.
\item The world's best soccer teams and popular players including Cristiano Ronaldo of Portugal and Lionel Messi of Argentina participated in the competition.
\end{itemize}

Sample C:

\begin{itemize}
\item The disaster occurred in Oman, specifically near Muscat, while also affecting surrounding countries like the United Arab Emirates and Iran.
\item The wildfires occurred in Australia, with New South Wales being the most affected state and major cities like Melbourne and Sydney also impacted.
\item The disaster occurred in South Korea, hitting several southern industrial hubs and a residential area in the southeastern port city of Pohang.
\item The disaster occurred in North and South Korea, with notable impacts being reported in Busan, South Korea's second-largest city, as well as Okinawa, the Ryukyu Islands, and Jeju Island.
\item The disaster occurred in an office building near the District Court in the South Korean city of Daegu.
\end{itemize}
\section{Human Annotations and Instructions}
\label{a:i}

We crowdsource human annotators for the MultiVENT task via Amazon Mechanical Turk to assess human performance and confidence. We first select annotators to perform the task by assessing their performance on a pilot task in which they write a natural language hypothesis about a news video, and rate their confidence that the hypothesis is true. We additionally provide a bot-checking question that asks annotators to answer a question differently if they are not a human (this filtered multiple annotators). The remaining pilot annotations were hand-evaluated by authors and the top scorers were allowed to participate on the full task. Annotators were paid at approximately \$15 USD per hour. Annotation instructions for the full task can be found below.

\subsection{Instructions}
\prompt{Human Annotation Instructions}{
In this task, you are presented with a video.
Please watch the video in full, with sound. Then, select the statement that you think is most likely to be true about the video content.
After selecting a statement, rate how likely you think it is that you are correct, on a scale from 0\% to 100\%.\\

Use the following rating scale:\\

\textbf{0\% (virtually impossible):} Essentially no way the hypothesis could possibly be true, given the evidence. Less likely than being struck by lightning.

\textbf{10\% (unlikely):} The hypothesis is unlikely, but definitely not impossible.

\textbf{20\% (possible): }The hypothesis could be true given the evidence, but there is better chance that it is false. Less likely than drawing a card of the suit of clubs from a standard card deck.

\textbf{30\% (reasonable chance):} You would not be more than mildly surprised that the hypothesis is true. About one thirds chance.

\textbf{40\% (a bit less than even-odds):} Slightly below fifty-fifty probability. You would not bet more than a small sum that the hypothesis is false.

\textbf{50\% (fifty-fifty):} Given the information about the situation, there is approximately equal chance that the hypothesis is true vs. the hypothesis is false. As likely as a fair coin landing on heads.

\textbf{60\% (a bit more than even-odds):} Slightly above fifty-fifty probability. You would not bet more than a small sum that the hypothesis is true.

\textbf{70\% (probable):} Likely, but you would still not be overly surprised if the hypothesis turned out to be false.

\textbf{80\% (quite likely):} About as likely as \*not\* rolling a “2” with a six-sided die.

\textbf{90\% (extremely likely):} Quite certain. You would bet a large amount of money on the hypothesis being true.

\textbf{100\% (practically certain):} You cannot imagine a scenario in which the hypothesis is not true, given the situational evidence.\\

Thank you for participating in the task. Quality of submitted tasks will be carefully monitored, and bot outputs will be rejected.
}

\subsection{Hypothesis Generation}
For hypothesis generation, news articles corresponding to the MultiVENT English current events were retrieved by hand using Google search. These were cleaned with the Newspaper3k Python package, and passed into GPT-4 alongside instructions. 
%We structure hypothesis generation around the prompts used for template extraction in \citet{sanders2024grounding}. 
The full prompt can be found below.

\prompt{Hypothesis Generation (Instructions, MultiVENT)}{
You are a summarization bot that takes a news article and a set of questions about it, and for each question outputs a sentence about the article that answers it. The answer should be independent and not reference the question or other answers. If the article does not provide an answer to the question, respond with "No answer". Otherwise, provide complete and detailed sentences.\\
Do not mention the article - your answers should be stand-alone statements about the article's contents.\\

\{Exemplar\}\\

CONTEXT: {context}\\

ARTICLE:\\
```\\
{article}\\
```\\

QUESTIONS:\\
{questions}\\

DETAILED ANSWERS (OR "NO ANSWER" IF APPLICABLE):
}

\prompt{Hypothesis Generation (Exemplar, MultiVENT)}{
CONTEXT: The following is an old article about a disaster that occurred.\\

ARTICLE:\\
```\\
Townsville floods: dam management under scrutiny as experts demand greater preparation\\

Study rates this week’s unprecedented monsoonal rainfall as comparable to a one in 1,000-year flood\\

Australian authorities must plan for more severe floods, experts have warned, amid concern that the management of the Townsville Ross River dam contributed to extensive flooding when a massive amount of water was released downstream.\\

The decision to open the floodgates when the dam water level reached 43m was based on a 2012 study that warned doing so could cause more extensive property damage during the most extreme flooding events.\\

The dam’s spillway gates were fully opened late on Sunday night, just before the dam level peaked at 42.97m, and in line with a current management plan. About 1,900 cubic metres of water a second gushed into the Ross River and ultimately into low-lying Townsville suburbs, where residents had to be evacuated.\\

Two fatalities have been reported after bodies were found in floodwaters. Emergency workers and the army said they had received more than 1,000 calls for help. They've been using boats and helicopters to move people to...\\
```\\

QUESTIONS:\\
(1) What disaster occurred?\\
(2) Where did the disaster occur?\\
(3) When did the disaster occur?\\
(4) What was the outcome of the disaster?\\
(5) Who was affected by the disaster?\\
(6) Who responded to assist those affected by the disaster?\\

DETAILED ANSWERS (OR "NO ANSWER" IF APPLICABLE):\\
(1) A severe flood occurred, due to monsoonal rainfall and the opening of the Ross River dam floodgates, which resulted in a massive amount of water being released downstream.\\
(2) The Ross River and downstream low-lying suburbs were flooded in Townsville, Australia.\\
(3) No answer.\\
(4) Emergency services received over 1,000 calls for help, and two fatalities were reported.\\
(5) Townsville suburb residents had to be evacuated, and bodies were found in floodwaters.\\
(6) Emergency workers and the army have been using boats and helicopters to move people.\\
}
\section{Other Prompts}
\label{a:p}
\subsection{Sample of reasoning prompts}
\prompt{Decomposition}{
Your task is to break down the following statement into two, simpler sentences. The sentences should be understandable on their own (e.g., no pronouns referring to the other sentence). If the statement can't be broken down further, only write "N/A".\\

STATEMENT: "Jason asked about the brown briefcase because he was concerned that it had been misplaced or stolen."\\
DECOMPOSITION:\\
(1) "Jason asked about the brown briefcase."\\
(2) "Jason was concerned that the brown briefcase had been misplaced or stolen."\\

STATEMENT: "{statement}"\\
DECOMPOSITION:
}
\prompt{Transcript Evidence Extraction (Offline)}{
You are a fact-checking expert that uses evidence to answer questions about a video.\\

For the following question and video dialogue transcript, write a set of up to three independent inferences entailed by some part of the scene. The inferences should resemble short, factual statements about the scene and should help to answer the question using component reasoning steps. The inferences should be understandable on their own (e.g., no pronouns or referring expressions that referring to other inferences).\\
Write your inferences in list format, i.e. \\
(1) Inference \#1\\
(2) Inference \#2\\
(3) Inference \#3\\
Write nothing other than your inferences.\\

QUESTION: "\{question\}"\\

TRANSCRIPT:\\
\{dialogue\}\\

INFERENCES (up to 3 total):
}
\prompt{Video Evidence Extraction (Offline)}{
You are a fact-checking expert that uses visual evidence to help a human answer questions about a video.\\

Given the video screenshot and following question, write a set of up to three independent inferences entailed by some part of the image. The inferences should resemble short, factual statements about the scene and should help to answer the question using component reasoning steps. The inferences should be understandable on their own (e.g., no pronouns or referring expressions that refer to other inferences). There may be names in the question, but you NEVER use people's names in your inferences. Instead of names, you describe people using common nouns like "man", "woman", or "person". For example, instead of writing "Ryan walks into a building", write "A man in a red shirt walks into a building".\\

Write your inferences in list format, i.e. \\
(1) Inference \#1\\
(2) Inference \#2\\
(3) Inference \#3\\
Write nothing other than your inferences.\\

If there is not relevant evidence in the image, only write "N/A". Do not write inferences about the question that do not pertain to the screenshot content. None of your inferences should directly reference the question. If there is no relevant visual content, write N/A.\\

QUESTION: "\{question\}"\\

INFERENCES (up to 3 total, not using any people's names) or N/A:
}

\prompt{Video Evidence Extraction (Test Time)}{
You are a fact-checking expert that uses visual evidence to help a human verify possibly false claims.
Given the following image-caption pair and a claim of unknown correctness, write a set of up to three independent inferences entailed by some part of the image. The inferences should resemble factual statements about the image and should help to determine whether the claim is true or false using component reasoning steps. The inferences should be understandable on their own (e.g., no pronouns or referring expressions that refer to other inferences). You should cite specific numbers listed in the image, if applicable.\\

Write your inferences in list format, i.e.\\
(1) Inference \#1\\
(2) Inference \#2\\
(3) Inference \#3\\
Write nothing other than your inferences.\\

If there is not relevant evidence in the image-caption pair, only write "N/A". None of your inferences should directly reference the question. If there is no relevant visual content, write N/A.\\

CLAIM: "\{question\}"\\

INFERENCES (max 3, or N/A):
}

\end{document}